# A Federated Artificial Intelligence Framework for Optimizing Pancreatic Cancer Treatment - Strategy Update

Anne-Christin HAUSCHILD [a,b,c,1], Amirreza ALEYASIN[a], Nils H. BEYER[a,d], Lisa FRICKE[e,f], Jonas HÜGEL[a,b,g], Maryam MORADPOUR[a,b,c], Anh-Tien NGUYEN[a,b,c], Youngjun PARK [a,b,h], Sophia RHEINLÄNDER[a,d], Tim BEISSBARTH[b,d,i,], Elisabeth HESSMANN[d,j], Martin MIDDEKE[k], Matthias LAUTH[l], Maximilian REICHERT[d,e,m-p], and Ulrich SAX [a,,b,d]

*[a] University Medical Center Göttingen, Department of Medical Informatics, Göttingen, Germany;*
*[b] University of Göttingen, Campus Institute Data Science (CIDAS), Section of Medical Data Science, Göttingen, Germany;*
*[c] Institute for Predictive Deep Learning for Medicine and Healthcare, Justus-Liebig University, Gießen, Germany;*
*[d] University Medical Center Göttingen, Clinical Research Group 5002 (CRU5002), Göttingen, Germany;*
*[e] Translational Pancreatic Cancer Research Center, TUM School of Medicine and Health, Department of Clinical Medicine – Clinical Department for Internal Medicine II, TUM University Hospital, Technical University of Munich, Munich, Germany;*
*[f] TUM School of Medicine and Health, Department of Clinical Medicine – Clinical Department for Internal Medicine II, University Medical Center, Technical University of Munich, Munich, Germany;*
*[g] Department of Health Technology, Technical University of Denmark, Kongens Lyngby, Capital Region, Denmark;*
*[h] Max Planck Institute for Biology of Ageing, Köln, Germany;*
*[i] University Medical Center Göttingen, Department of Medical Bioinformatics, Göttingen, Germany;*
*[j] University Medical Center Göttingen, Department of Gastroenterology, Gastrointestinal Oncology and Endocrinology, Göttingen, Germany;*
*[k] Philipps-University Marburg, Comprehensive Cancer Center, Marburg, Germany;*
*[l] Philipps-University Marburg, Clinic for Gastroenterology, Endocrinology and Metabolism, Marburg, Germany;*
*[m] Center for Organoid Systems (COS), Technical University Munich (TUM), 85747 Garching, Germany;*
*[n] German Cancer Consortium (DKTK), partner site Munich, a partnership between DKFZ and University Hospital Klinikum rechts der Isar, Munich, Germany;*
*[o] Bavarian Cancer Research Center (BZKF), Munich, Germany*

**Abstract. Introduction**: While a centralized approach involving patient consent to collect and analyze data centrally would theoretically offer the best data quality and predictive performance, it is not always feasible in practice. Federated Learning (FL) or Federated Artificial Intelligence architectures have shown to be a very promising approach to use and access distributed disease related resources within the GDPR boundaries. In a previous case report, we describe the preconditions at the participating sites, the necessary administrative and process related steps to prepare data, people and infrastructure for improving subtype identification and assessing treatment options in pancreatic cancer. We update this case report sharing our experience tackling the challenges and show some preliminary results of the actual federated learning AI pipelines in our pancreatic cancer projects. **Methods**: Starting at the participating sites, we have to identify and to annotate the data being accessible after extraction and transformation in a local FL hub - in our case a centrally developed and distributively deployed docker container. This docker container comprises the FL scripts generating local models. We apply a newly developed FL algorithm considering all local features, including partial overlapping features specific to the local sites. **Results**: Theoretically, an annotation in a cancer setting should succeed using the German oncology core data set (oBDS), which is already utilized for mandatory reporting to cancer registries, and can be sustained in the FL setting. The FL algorithms deal robustly with partially overlapping features as we showed with public data sets. **Discussion**: Major roadblocks including straightening the operational concepts for the infrastructures, the ethics approval for such novel architectures and support for every site have been addressed. However, scaling up this approach in the future faces hurdles; while including broader multi-modal data sets should be feasible, large-scale deployment to many more sites remains challenging.

[1]Corresponding author: Anne-Christin Hauschild, Justus-Liebig University Gießen, Institute for Predictive Deep Learning for Medicine and Healthcare, Friedrichstraße 25, 35392 Gießen, Germany; E-Mail: anne-christin.hauschild@uni-giessen.de

**Keywords.** Federated Learning, Therapeutics, Pancreatic Cancer, Data Management

## 1. Introduction

### *1.1. Background*

Despite significant scientific and clinical efforts, the aggressive tumor biology and remarkable resistance to conventional anti-tumor treatments solely led to slow improvements of 5-year PDAC survival rates (not more than 10%) compared to other cancer modalities [1]. While predictive Artificial Intelligence (AI) has achieved remarkable progress, revolutionizing medical research and applications where well-structured BIG data is available, we lack infrastructure for learning in scarce distributed data situations, such as for rare and heterogeneous diseases that require personalized treatment approaches. Additionally, the General Data Protection Regulation (GDPR) regulates data privacy and restricts the sharing of data within Europe [2].

This fueled the growing interest in novel privacy-preserving federated learning (FL) AI architectures and infrastructures, particularly in the healthcare domain. FL allows training models without the need for data centralization by only exchanging models or their parameters to build a centrally combined learning model [5]. However, many challenges arise when building suitable federated infrastructures and AI models; for instance, (1) expensive communication, (2) heterogeneous system, (3) data heterogeneity, and (4) privacy [3].

The “FAIrPaCT” research network (Federated Artificial Intelligence Framework for Optimizing Pancreatic Cancer Treatment), funded by the BMFTR under the National Decade Against Cancer initiative, aims to develop a federated AI software system to predict the success of individual pancreatic cancer treatments and identify factors that improve therapy effectiveness. The network aims to harness the unique and extensive clinical, molecular and imaging data from Germany’s largest PDAC patient cohorts, located at major German medical centers, including the University Medical Center Göttingen, the University Hospital of Gießen and Marburg, and the Rechts der Isar Hospital of the Technical University of Munich.

### *Objective and Requirements*

The main objective of the FAIrPACT project is to estimate individual patient prognosis, survival, and treatment success in pancreatic cancer. To fulfill this objective, large case numbers are required that usually cannot be reached in only one location. Therefore, in this study the data is acquired from three different locations, allowing for federated analysis beyond the boundaries of one location. In the future, the system will be usable regardless of location. In addition, it aims to identify important parameters that influence the patient trajectory and response to a specific treatment. By using the data collection from three locations, the project aims to find general factors for the therapy success against pancreatic cancer. Which would represent an important milestone towards precision medicine supported by artificial intelligence.

In the long term, researchers, patients with pancreatic cancer, but also oncologists, should benefit from the enlargement of the dataset. The analyzed data might help identify important clinical determinants for treatment success, and build more reliable models to support treatment decisions. In addition, the “FAIrPaCT” research network aims to achieve a better understanding of the tumor evolution and progression in pancreas and thus identify the underlying molecular mechanisms that lead to the success or failure of a treatment. On the one hand, this can pave the way for the development of new personalized therapies. On the other hand, we aim to develop a prove-of-concept for the use of federated AI based software systems supporting both science and clinical practice. Most importantly, FAIrPaCT aims to develop reusable, reliable open-source framework comprising federated learning algorithms and prototypes ready for deployment towards other oncological diseases in the future.

Main Objectives:

1. Identification, standardization and homogenization of PDAC data according to FAIR criteria (Findable Accessible Interoperable Reusable) and development of a standardized PDAC information model and data management for heterogeneous, and partially overlapping clinical, omics and imaging data.
2. Development and deployment of robust Federated Artificial Intelligence (FAI) models for PDAC treatment prediction on distributed partially overlapping non-IID biomedical and imaging data.

## 2. State of the art

### *2.1. Related Work*

Previously, different federated frameworks have been developed and employed focusing on sensitive areas like healthcare:

The DataSHIELD (Data Aggregation Through Anonymous Summary-statistics from Harmonised Individual levEL Databases) infrastructure represents another innovative approach to secure biomedical data analysis that addresses privacy and ethical concerns in collaborative research [6]. This federated system enables researchers to perform statistical analyses on sensitive individual-level data across multiple studies without requiring physical pooling of the raw data, operating on a client-server model where analysis requests are sent to data repositories where computations occur behind secure firewalls [7]. DataSHIELD supports various analytical functions including descriptive statistics, contingency tables, and generalized linear models while maintaining data privacy through multiple safeguards such as restricted function access and disclosure controls. Recent developments have expanded DataSHIELD's functionality to include applications in structured text analysis, and advanced data visualization techniques, while ongoing work in the community focuses on mapping processes that facilitate interoperability between different data systems and standards [8]. DataSHIELD is intensively used, e.g. in the nutrition epidemiology since many years [9].

Finally, another approach of federated analyses uses an interesting “train” paradigm. The Personal Health Train (PHT) focuses on complementing the NFDI4Health data infrastructure with analytics considering the distribution of data collections. NFDI4Health aims to develop a data infrastructure for personalized medicine and health research and to make data generated in clinical trials, epidemiological, and public health studies FAIR (Findable, Accessible, Interoperable, and Reusable) [10].
The FeatureCloud project aims to lower the technical challenges of using federated learning by providing a user-friendly platform supporting privacy-preserving machine learning between multiple institutions [4]. The decentralized design follows strict data protection rules by keeping the raw data on local servers and sharing updated model parameters between sites. A major benefit of the FeatureCloud platform is the support of a built-in AI store, incorporating dockerized workflows to allow easy access and setup in various existing IT systems. This includes preprocessing utilities like normalization, one-hot encoding or dimensionality reduction, as well as AI methods such as Logistic Regression, Random Forest, or Deep Learning for various applications such as survival analysis. Also, specialized tools for tasks like GWAS based on PLINK are available. This helps developers to share, publish, and use federated learning algorithms and encourages collaboration and research in the field [4].

### *2.2. Shortcomings*

While previous frameworks support various aspects of federated analysis, they face a multitude of challenges and we can observe a significant delta between the described requirements and the solutions available to date. For example, the DataSHIELD approach has worked well in nutritional epidemiology for many years, as the involved partners are disciplined and experienced in transforming their harmonized datasets into the required OPAL database to enable access for external analysis logic. These ideal conditions may not be given in many other loosely coupled projects and even problematic in only partially overlapping datasets [7]. FeatureCloud excels as a robust and adaptable tool, facilitating secure and private collaborative data analysis and AI, which aids in resolving critical issues in medical AI. However, the implementation of FeatureCloud in clinical settings presents challenges such as the need for stable network connections, handling diverse datasets, and navigating complex data protection laws [4]. The Personal Health Train supports federated data analysis and its interoperability has already been tested in a pilot study of two PHT infrastructures at three different locations [10]. Previous systems support a variety of data science tools, statistical analysis and even AI, but major challenges remain. In the following, we mention some of the most important issues:

- Local accessibility and heterogeneity of the data in particular when including multi-omics data modalities that are partially location specific.
- The quality of the data regarding missing values, divergence in encoding or varying recording standards in terms of frequency or content, just to name a few.
- Local infrastructure, data interoperability as well as availability of computational resources required for training specific federated AI models.
- Organizational challenges.

## 3. Concept

The FAIrPaCT framework architecture consists of Docker containers deployed at the partner sites Göttingen, Marburg and Munich, ensuring local data privacy while facilitating global model development. Each site independently preprocesses its data, trains local models and evaluates performance while contributing to a shared global model. At each site, Docker 0 handles the Extract, Transform, Load (ETL) process to harmonize the local dataset. This step ensures high-quality preprocessing of pancreatic cancer data to maintain consistency and reliability across sites. Once preprocessed, the data is split into training and test subsets. Docker 1 utilizes 80% of the training data to train machine learning models locally. These locally trained models are then transferred to Docker 2 via the File Transfer Application Platform for Integration (FTAPI) service of the University Medical Center Göttingen. Docker 2 acts as the central coordinator for federated learning. Docker 2 decrypts the local models received from each site and aggregates them into a single global model. This aggregation process preserves data privacy while leveraging insights from diverse datasets across partner sites. The global model is subsequently sent back to each site via FTAPI.
Docker 3 at each site evaluates both the local and global models using the remaining 20% test data. This evaluation step involves performance comparisons to assess the robustness and accuracy of the global model in relation to locally trained models. By enabling this iterative process, the framework ensures that insights from individual datasets contribute to a more generalizable and robust AI model.

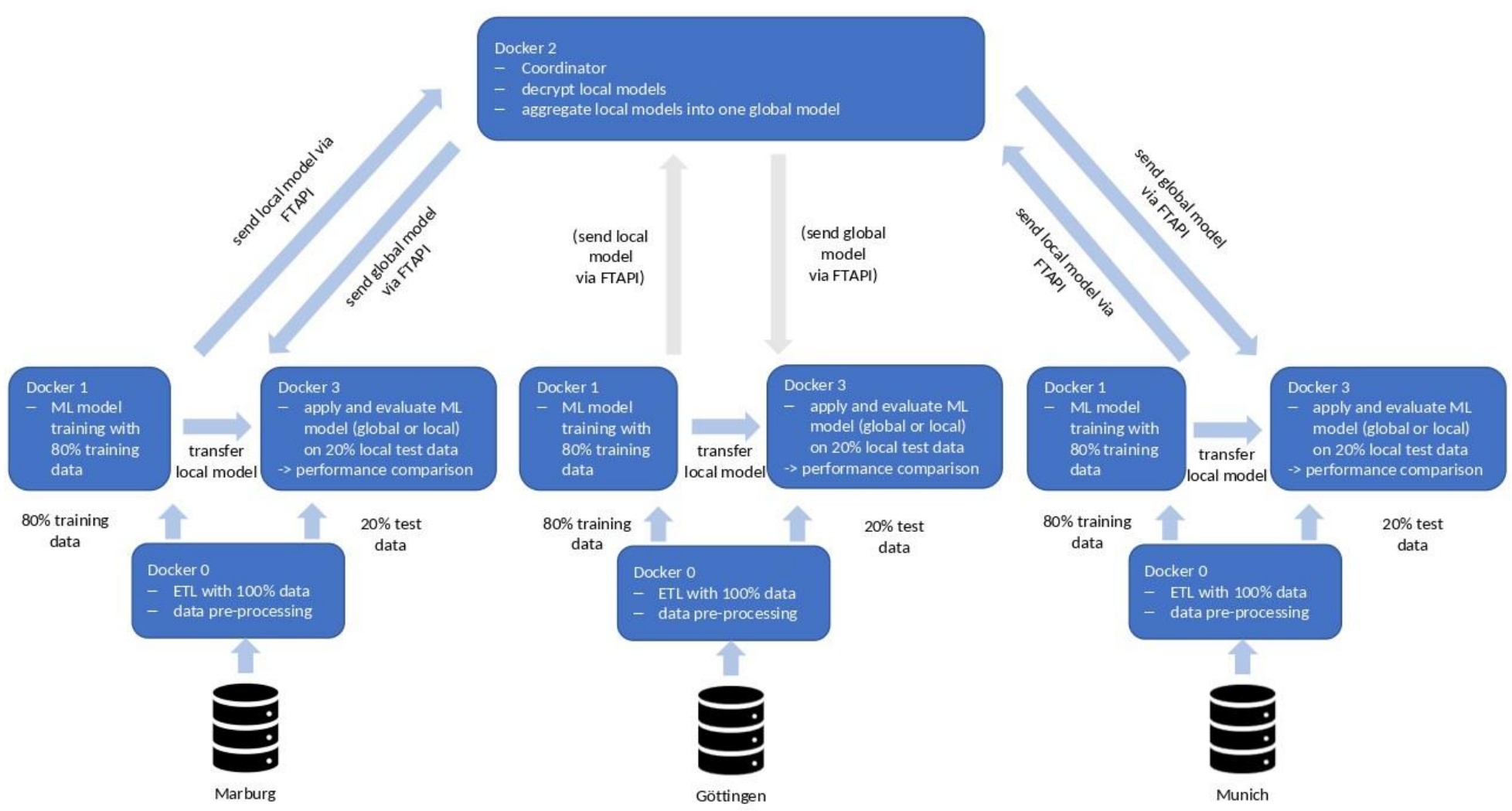

**Figure 1. Architecture of the FAIrPaCT framework**.

Prior to the Docker-based preprocessing stage, clinical data is extracted from ONKOSTAR, a tumor documentation system deployed at the University Medical Center Göttingen. ONKOSTAR operates within the so called *Patlan*, a dedicated, access-controlled network segment designed to meet stringent data protection requirements for sensitive patient information. Nevertheless, the model training and most data preprocessing should not take place in these protected network segments. To make pseudonymized patient data available and apply ETL scripts, it is essential to identify a less restricted network segment that is accessible from the *Patlan* and supports dedicated external connections. It is even more crucial to request servers and firewall rules ahead of time to implement ETL methods to access the data in ONKOSTAR while avoiding unnecessary delay in the project. While ONKOSTAR offers flexibility in clinical documentation, its heterogeneous data structures pose challenges for federated machine learning across sites. To ensure algorithmic interoperability, the framework mandates compliance with the standardized core oncology dataset oBDS. The oBDS extraction from ONKOSTAR requires bypassing GUI and REST API limitations via direct SQL queries to the underlying database. A custom ETL script retrieves tumor-specific reports while preserving ontological relationships between clinical entities. Prior to transfer to the research environment, data undergoes pseudonymization and sensitivity filtering. Pseudonymization involves replacing direct identifiers with reversible hashes, and sensitivity filtering involves rule-based redaction of free-text fields containing potential re-identification vectors.
The sanitized oBDS records are securely transferred to a research data repository within the Göttingen network infrastructure. Docker 0 orchestrates the unification of three data modalities: clinical data in the form of processed oBDS records, genomic data including panel sequencing in Variant Call Format (VCF) files, and RNA sequencing data in the form of count files. Cross-modal linkage occurs through pseudonymized patient IDs. The oBDS - centric approach emerged from practical constraints: developing preprocessing pipelines without direct data access due to privacy regulations necessitated reliance on legislatively mandated standardization. While this ensures cross-site consistency, it introduces limitations in leveraging site-specific documentation nuances, a trade-off critical for enabling privacy-preserving federated learning.

## 4. Implementation

### *4.1. Challenges and Solutions / Results*

One of the primary challenges in the context of handling pancreatic cancer data across multiple partner sites is the data harmonization. Data heterogeneity complicates efforts to integrate data into a cohesive framework suitable for machine learning and artificial intelligence applications. Addressing these issues requires robust preprocessing pipelines and alignment of metadata standards to ensure that data quality and interoperability are maintained.
A specific example of this challenge is the export of clinical patient data from the local tumor documentation systems at the partner sites. The oBDS was identified as suitable because it serves as a standardized dataset for oncology documentation in Germany, designed to facilitate reporting to cancer registries and support clinical research [11]. While ONKOSTAR, which is used in Göttingen and Marburg, provides interfaces for exporting oBDS data in formats such as XML, integrating these exports into broader frameworks like the FAIrPaCT system requires additional steps. These include pseudonymization, de-identification, and ensuring that the exported data aligns with the requirements of federated learning models.
To address the challenge of heterogeneous datasets that only share specific sets of parameters across different sides, the FAIrPaCT team developed the FRF4POD (Federated Random Forest for Partially Overlapping Data) algorithm. FRF4POD is based on the traditional Random Forest and addresses the problem of data heterogeneity. Many federated learning methods train one shared federated model. In contrast, FRF4POD builds a side-specific model that enables the use of the entire local dataset. It does this by choosing matching decision trees from the shared global forest. Each site uses only the trees that match its own features. This way, each institution can improve its predictions while keeping the model closely linked to its local clinical data. FRF4POD is available as an open-source Python package. Overall, FRF4POD is a strong, scalable, and privacy-friendly solution. It is especially useful for collaborative machine learning in healthcare and other sensitive fields [12].

In two related projects, the MTB-Report and the CRU 5002, we already developed ETL pipelines to integrate the patient data exported from ONKOSTAR into other tools such as cBioPortal or tranSMART [13]. While the end systems of these pipelines differs compared to those in FAIrPaCT, considerations regarding the network segments are similar. By building on our experience and infrastructure from implementing these pipelines, we limited the required new firewall rules and severs, therefore speeding up the data access in Göttingen location.

### *4.2. System in Use*

At the University Medical Center Göttingen (UMG), RNA-seq data were available for Patient-Derived Xenograft (PDX), Cell-Derived Xenograft (CDX), and organoid samples, whereas the Munich RNA-seq dataset comprised organoid samples. Comparison of the genes represented in the RNA-seq datasets from UMG and Munich identified 12,695 overlapping genes.

**Table 1. Overview of available data modalities and quantities within the FAIrPaCT consortium**

| | Göttingen | Marburg | Munich |
|---|---|---|---|
| Clinical Features (Patients) | 181 (104) | 60 (183) | 539 |
| Panel-Seq Genes (Patients) | 98 (60) | No Panel-Seq | 465 |
| Panel-Seq SNPs | 7384 | No Panel-Seq | >900 |
| H&E Imaging | 517 | 183 | 197 |

In addition, 7,384 unique SNPs were identified at UMG, spanning 98 genes across 60 patients. (See Table 1). Due to the limited pseudonymized sample set provided by TUM, the total number of SNPs across the full cohort in Munich remains uncertain at UMG's site, see Figure 2.

To further analyze the 7,384 SNPs identified in UMG patients, a threshold was applied to retain only SNPs occurring in at least two patients. This reduced the total number of SNPs from 7,384 to 2,236, revealing that approximately 5,100 SNPs are unique to individual patients. Increasing the threshold further continues to refine the set of shared mutations across patients.

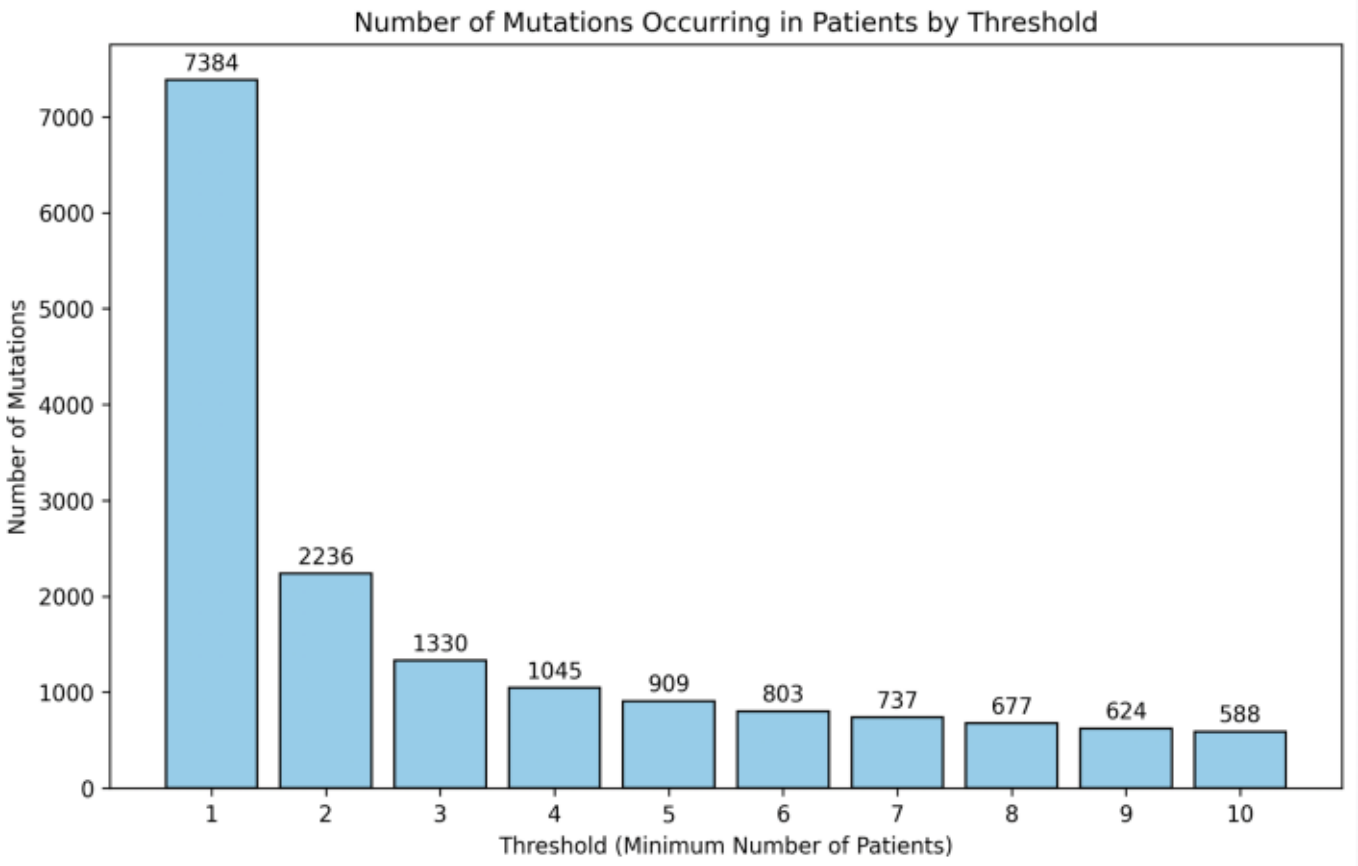

**Figure 2.** Distribution of mutations based on the number of patients in which they occur at UMG.

As part of the ongoing implementation, the FAIrPaCT Docker container has been successfully deployed and executed at the Marburg site. This marks a significant milestone in the project, as it enables the local preprocessing and training of models on Marburg's clinical and genomic data and was made possible due to the standardization of the oBDS-format. Initial statistics have been generated from the Marburg dataset, including descriptive summaries of clinical features. These statistics are currently being compared with those from Göttingen and Munich to assess data consistency and identify potential harmonization needs. The successful container deployment in Marburg demonstrates the portability and robustness of the FAIrPaCT framework across different institutional IT environments.

## 5. Lessons learned

The implementation of the FAIrPaCT framework revealed several crucial insights for future multi-site federated learning projects in healthcare:

1. Technical Liaison Importance: A significant challenge was the lack of dedicated technical liaisons at the participating sites. Such roles are essential for implementing algorithms, deploying Docker infrastructure, and iteratively resolving issues that only become apparent during initial implementations. Future projects should ensure each site has a designated technical expert with both the knowledge and resources to actively engage in cross-site collaboration.
2. Clear Responsibility Distribution: The project would have benefited from a more explicit delineation of responsibilities from its inception. This includes not only technical roles but also project management and communication channels.
3. Communication Protocols: Effective inter-site communication emerged as a critical factor. Establishing robust communication protocols and regular check-ins could mitigate issues such as task duplication or oversight.
4. Early Bottleneck Identification: A more proactive approach to identifying potential bottlenecks in the workflow could have streamlined the project timeline. This includes both technical and administrative hurdles.

5. Ethics Approval Streamlining: Obtaining ethics approval proved to be a significant delay factor, as data availability was contingent on this process. Future initiatives should consider parallel processing of ethics applications or explore ways to expedite this critical step.
6. Data Standardization Challenges: One site's non-standardized clinical data format highlighted the importance of establishing common data models early in the project. The initial decision to abandon standardization efforts, followed by their subsequent revival due to resource constraints, underscores the need for comprehensive planning in data harmonization.
7. Holistic Standardization Approach: Beyond data format standardization, the project revealed the necessity of standardizing the entire data generation pipeline, particularly for genomic data such as panel and RNA sequencing. This ensures consistency in data quality and interpretability across sites. Nevertheless, the same applies for the used clinical data. In particular, the data preparation (including the data cleaning) should be harmonized to extract meaningful features from secondary use routine ONKOSTAR data (compare with [14]).
8. Resource Allocation: The project underscored the importance of adequate resource allocation, both in terms of personnel and funding. Ensuring that key stakeholders have the necessary incentives and support to fully engage with the project is crucial for its success.
9. Server and Network selection/planning: Identifying the correct network segments and setup the required servers in time, requires in-depth knowledge of the available infrastructure, as well as (local) regularization regarding firewall rules and accessibility of the servers for the ETL scripts and the ML. A direct contact within the IT support / server center would be beneficial to plan ahead and avoid unnecessary delays through long conversations and discussions where to place and how to access servers.

These lessons provide valuable insights for improving the design and execution of future federated learning projects in healthcare, emphasizing the need for robust planning, clear communication, and standardized processes across all participating sites.

## 6. Conclusion

Federated learning offers a promising approach to collaborate even with partially overlapping data sets in different institutions. As compared to other federated analysis approaches FL can still deliver results.
Nevertheless, this approach entails trade-offs, underscoring that a centralized solution would be more advantageous whenever feasible, particularly in prospective settings.

## Declarations

*Ethical vote:*
Ethical committee: Ethik-Kommission der Universitätsmedizin Göttingen, chair: Prof. Dr. Jürgen Brockmüller, vote-no:24/2/24, date: 9.10.2024
The authors declare, that there is no conflict of interest.

*Contributions of the authors:*
All authors were involved in the planning and implementation of the project. US did plan, choose format and organize. US, NHB and EH prepared the ethics vote. ACH, NHB, LF, SR, MR, JH, and US were involved in writing and substantial revising of the manuscript.
*All authors critically read, reviewed and approved the manuscript*.

*Acknowledgement:*
We are very grateful for the funding of our research projects.
Funding from the BMFTR in the FAIrPaCT projects (BMFTR 01KD2208A, 01KD2414A) and by the Innovation Committee at the Federal Joint Committee NO. 01VSF20014 (KI-Thrust).
Funding from the DFG in the project KFO5002 (DFG 426671079), KFO325 (DFG 329116008), NFDI4Health (DFG 442326535). Partially funded by the EU in SHARE-CTD (HORIZON-MSCA-2022-DN 101120360).